\pdfoutput=1
\documentclass[runningheads]{llncs}

\usepackage{url}

\usepackage{amsmath}
\usepackage{amssymb}

\usepackage{cellspace}

\usepackage{pgfplots}

\usepackage{mathtools, nccmath}

\usepackage{tikz}
\usetikzlibrary{shapes}

\usepackage{algorithm}
\usepackage{algpseudocode}

\tikzset{
circ/.style={draw,circle,minimum height=3em},
}

\begin{document}
\setcounter{MaxMatrixCols}{50}

\title{Conceptual Evaluation of Deep Visual Stereo Odometry for the MARWIN Radiation Monitoring Robot in Accelerator Tunnels}

\author{André Dehne \and Juri Zach \and Peer Stelldinger\\
\institute{Faculty of Computer Science and Digital Society, HAW Hamburg, Germany} \email{\{andre.dehne, juri.zach, peer.stelldinger\}@haw-hamburg.de}
}
\authorrunning{A. Dehne \and J. Zach \and P. Stelldinger}

\maketitle              

\begin{abstract}
The MARWIN robot operates at the European XFEL to perform autonomous radiation monitoring in long, monotonous accelerator tunnels where conventional localization approaches struggle. Its current navigation concept combines lidar-based edge detection, wheel/lidar odometry with periodic QR-code referencing, and fuzzy control of wall distance, rotation, and longitudinal position. While robust in predefined sections, this design lacks flexibility for unknown geometries and obstacles. This paper explores deep visual stereo odometry (DVSO) with 3D-geometric constraints as a focused alternative. DVSO is purely vision-based, leveraging stereo disparity, optical flow, and self-supervised learning to jointly estimate depth and ego-motion without labeled data. For global consistency, DVSO can subsequently be fused with absolute references (e.g., landmarks) or other sensors. We provide a conceptual evaluation for accelerator tunnel environments, using the European XFEL as a case study. Expected benefits include reduced scale drift via stereo, low-cost sensing, and scalable data collection, while challenges remain in low-texture surfaces, lighting variability, computational load, and robustness under radiation. The paper defines a research agenda toward enabling MARWIN to navigate more autonomously in constrained, safety-critical infrastructures.
\end{abstract}


\section{Introduction} \label{deh:introduction}

Autonomous robots operating in radiation-exposed environments face unique challenges that combine safety-critical navigation with harsh operating conditions. The European X-ray Free-Electron Laser (European XFEL) facility exemplifies such an environment, where autonomous radiation monitoring is essential to maximize accelerator uptime while minimizing human exposure to hazardous conditions~\cite{DIO2022}. With the XFEL generating high-energy X-ray flashes and operating along kilometer-long tunnel sections, manual radiation inspections are time-consuming and pose risks to personnel safety. The MARWIN (Mobile Autonomous Robot for Maintenance and Inspection) platform was developed to address these challenges through autonomous radiation measurement in the European XFEL's accelerator tunnels~\cite{DEH2018}. Operating in GPS-denied environments characterized by straight, monotonous geometry with sections up to 1,800 meters in length, MARWIN must achieve centimeter-level positioning accuracy to safely approach measurement targets without collision. The current system relies on a multi-sensor approach combining 2D LIDAR scanners for obstacle detection and wall distance measurement, wheel odometry for relative positioning, and QR codes mounted at 10-meter intervals for absolute localization corrections~\cite{DEH2018}. While this hybrid approach has proven effective for structured measurement tasks in known tunnel sections, it exhibits limitations in flexibility and scalability. The dependency on pre-installed QR code infrastructure restricts deployment to prepared environments, while LIDAR-based localization struggles in sections with minimal geometric features or when facing dynamic changes to the tunnel layout. Furthermore, the reliance on static maps requires remapping efforts whenever tunnel configurations change due to maintenance or equipment modifications. Recent advances in deep learning-based visual odometry offer promising alternatives that could address these limitations. Deep visual stereo odometry (DVSO) systems combine stereo vision for metric depth estimation with self-supervised learning to jointly infer ego-motion and scene geometry without requiring labeled training data. Such approaches have demonstrated robust performance in challenging environments while offering potential advantages in cost, computational efficiency, and adaptability to novel scenarios. This paper presents a conceptual evaluation of DVSO as a complementary or alternative navigation approach for the MARWIN robot in accelerator tunnel environments. We analyze the specific requirements of radiation monitoring in the European XFEL context, examine the capabilities and limitations of current DVSO methods, and identify research directions toward enabling more flexible autonomous navigation in constrained, safety-critical infrastructures. Building upon the established operational experience with MARWIN~\cite{DEH2018} and considering the high-availability requirements of modern accelerator facilities~\cite{DIO2022}, we aim to chart a path toward next-generation robotic inspection systems that balance robustness, efficiency, and adaptability.

\section{Challenges in Accelerator Tunnels} \label{deh:challenges}

Accelerator tunnels present a uniquely challenging environment for autonomous navigation and simultaneous localization and mapping (SLAM) systems. The European XFEL facility \cite{DIO2022}, where the MARWIN robot operates \cite{DEH2018}, exemplifies these constraints through its kilometers-long underground tunnels with highly specialized operational requirements.

\begin{itemize}
    \item \textbf{Geometric Constraints:} The tunnel environment exhibits monotonous geometry with long, straight corridors that provide minimal loop-closure opportunities \cite{PRA2021}. Repetitive structures such as cable trays and equipment installations repeat at regular intervals, causing pose ambiguity for feature-based localization. These characteristics lead to cumulative drift with unbounded odometry error accumulation over multi-kilometer trajectories.

    \item \textbf{Perceptual Constraints:} Visual perception faces limited features from uniform, low-texture surfaces including concrete walls and metal housings, compounded by poor or uneven lighting conditions \cite{ZHA2022}. Semantic landmarks remain sparse, though component labels offer potential for self-supervised landmark detection.

    \item \textbf{Hardware Constraints:} Radiation tolerance requirements impose limited computational capacity through the use of radiation-hardened hardware systems \cite{DEH2018}. Battery-powered operation over extended inspection missions demands energy efficiency. Real-time processing requirements must balance accuracy with these limited processing capabilities \cite{FU2022}.

    \item \textbf{Environmental Constraints:} The underground setting provides no external absolute position references due to GPS-denial \cite{DEH2018}. Narrow passages impose spatial restrictions with limited maneuvering room for the mobile platform. Dynamic disturbances from electromagnetic interference generated by accelerator components can affect sensor reliability.
\end{itemize}

These challenges collectively violate fundamental assumptions underlying most SLAM research: feature-rich environments (KITTI, TUM-RGBD benchmarks provide abundant features), natural loop closures (urban/indoor scenarios offer revisitation), and high computational budgets (GPU or workstation resources). Addressing accelerator tunnel navigation requires specialized approaches that explicitly account for these extreme constraints.

\section{State of the Art} \label{deh:sota}

Building upon the challenges identified in Section~\ref{deh:challenges}, this section systematically analyzes existing approaches to odometry and SLAM, evaluating their applicability to the specific constraints of accelerator tunnel navigation. We organize the analysis around three focus areas (FA) that directly address the core challenges:

\begin{itemize}
    \item \textbf{FA1}: Self-supervised SLAM approaches in GPS-denied, monotonous tunnel environments (addressing challenges of repetitive geometry, lack of loop closures, and cumulative drift)
    \item \textbf{FA2}: Resource-efficient odometry and SLAM on embedded hardware (addressing radiation-hardened hardware constraints and real-time processing requirements)
    \item \textbf{FA3}: Self-supervised landmark discovery methods (addressing scarcity of semantic landmarks through automatic detection of component labels and infrastructure markers without requiring manual annotation)
\end{itemize}

Table~\ref{deh:literature_comparison} provides a systematic comparison of representative works across these focus areas, evaluating how each method addresses or fails to address the accelerator tunnel challenges.

\begin{table*}[htbp]
\centering
\caption{Overview of related SLAM and odometry work organized by focus areas}
\label{deh:literature_comparison}
\vspace{\baselineskip}
\scriptsize
\renewcommand{\arraystretch}{1.3}
\begin{tabular}{p{2.0cm}p{3.0cm}p{3.0cm}p{3.0cm}}
\hline
\textbf{Paper} & \textbf{FA1: Self-supervised SLAM in Tunnels} & \textbf{FA2: Resource-efficient SLAM} & \textbf{FA3: Self-supervised Landmark Detection} \\
\hline
Zhang et al. \cite{ZHA2022}
&
Classical RGB-D SLAM (Manhattan SLAM); Robust for tunnel geometry, no learning involved
&
Real-time on desktop hardware only; No embedded or energy efficiency evaluation
&
Uses geometric planes and lines only; No semantic or text landmarks
\\[0.5em]
\hline
Prados et al. \cite{PRA2021}
&
Graph-SLAM (ICP-based) in CERN SPS tunnel; Robust localization in GPS-denied environments
&
Computationally heavy; No optimization for limited hardware
&
Manual structural landmarks (doors, sections); No semantic or self-supervised detection
\\[0.5em]
\hline
Chen et al. \cite{CHE2023}
&
Fully self-supervised LiDAR mapping; Differentiable optimization without labels
&
Algorithmically efficient (no explicit loop-closure); GPU-dependent, not real-time embedded
&
No semantic or textual landmark integration
\\[0.5em]
\hline
Fu et al. \cite{FU2022}
&
Self-supervised LiDAR odometry (label-free); Uses geometric consistency losses
&
Real-time (20 Hz) on GPU; Efficient via 2D spherical projection
&
Purely geometric features; No semantic or label-based landmarks
\\
\hline
\end{tabular}
\end{table*}

\subsection{Analysis by Focus Area}

\subsubsection{FA1: Self-Supervised SLAM in Tunnels}

Tunnel environments pose unique challenges for SLAM due to GPS-denial, monotonous geometry, and limited loop-closure opportunities. Classical approaches like the MARWIN system \cite{DEH2018} rely on 2D LiDAR and wheel odometry with manually placed QR codes for drift correction, requiring infrastructure preparation. Zhang et al. \cite{ZHA2022} apply RGB-D SLAM with Manhattan world assumptions for tunnel reconstruction, achieving robustness through geometric constraints but remaining sensitive to lighting conditions. Prados et al. \cite{PRA2021} demonstrate graph-SLAM in the CERN SPS tunnel using ICP-based point cloud matching, successfully localizing in GPS-denied environments but experiencing significant drift in monotonous sections. Recent self-supervised approaches eliminate the need for labeled training data. Chen et al. \cite{CHE2023} present DeepMapping2, a fully self-supervised LiDAR mapping framework using differentiable optimization without explicit loop closure detection. Fu et al. \cite{FU2022} develop self-supervised LiDAR odometry based on geometric consistency losses and spherical projection. However, both methods focus on feature-rich outdoor scenarios and have not been validated in the extreme monotony of accelerator tunnels, where geometric features repeat identically over kilometer-long distances.

\subsubsection{FA2: Resource-Efficient SLAM}

The MARWIN robot operates on radiation-tolerant limited hardware \cite{DEH2018} with computational constraints below typical SLAM research platforms. Most learning-based SLAM systems assume GPU availability: Fu et al. \cite{FU2022} achieve 20 Hz real-time performance on GPU through efficient spherical projection, while Chen et al. \cite{CHE2023} require GPU acceleration despite algorithmic efficiency from eliminating explicit loop closures. Zhang et al. \cite{ZHA2022} evaluate their RGB-D SLAM only on desktop hardware without embedded deployment or energy consumption analysis. None of the surveyed works address the dual constraints of radiation tolerance and real-time embedded operation. The computational budget available on radiation-tolerant system designs precludes direct deployment of GPU-dependent methods, requiring fundamental algorithm redesign or hardware-aware optimization strategies not explored in current literature.

\subsubsection{FA3: Self-Supervised Landmark Discovery}

Accelerator tunnels lack natural semantic landmarks but contain infrastructure component labels and position markers that could serve as localization references. However, manual annotation is impractical given facility scale. While self-supervised learning has advanced in odometry and mapping, landmark discovery remains focused on geometric features. Chen et al. \cite{CHE2023} use self-supervised contrastive learning for loop closure detection based on geometric LiDAR patterns. Fu et al. \cite{FU2022} and Zhang et al. \cite{ZHA2022} similarly rely on geometric features (point clouds, planes, lines) rather than semantic content. Prados et al. \cite{PRA2021} incorporate manual structural landmarks like doors and tunnel sections, but these require human annotation and do not leverage the textual labels present on technical equipment. No existing work addresses self-supervised detection, localization, and encoding of text-based or symbolic infrastructure markers as stable landmarks for long-term navigation in feature-sparse environments.

\subsection{Identified Research Gap}

The systematic analysis reveals a critical gap when comparing existing methods against the challenges outlined in Section~\ref{deh:challenges}: \textbf{No existing approach jointly achieves resource-efficient odometry, robust operation in monotonous GPS-denied tunnels, and self-supervised landmark discovery.} 

Mapping the literature to the identified challenges shows:

\begin{itemize}
    \item \textbf{Monotonous geometry \& drift}: Classical methods (MARWIN \cite{DEH2018}, Prados et al. \cite{PRA2021}) address this through infrastructure (e.g., QR codes) or accept significant drift. Self-supervised methods have not been tested in such scenarios.
    
    \item \textbf{GPS-denial \& limited features}: While all surveyed methods operate without GPS, only tunnel-specific works explicitly address feature scarcity. Self-supervised landmark discovery methods focus on geometric features (Chen et al. \cite{CHE2023}) rather than semantic infrastructure markers.
    
    \item \textbf{Real-time processing \& energy efficiency}: DeepMapping2 \cite{CHE2023} requires hours of optimization; SSLO \cite{FU2022} needs GPU inference. No work evaluates inference latency and power consumption on embedded platforms.
\end{itemize}

This gap motivates the development of a comprehensive solution for autonomous navigation in accelerator tunnels that integrates robust odometry, self-supervised landmark learning, and resource-efficient implementation.

\subsection{Research Questions}

Based on the identified gap, the overarching research challenge is:

\textbf{How can a resource-efficient, self-supervised SLAM system be developed for autonomous robots operating in long, monotonous accelerator tunnels, capable of generating robust landmarks from automatically detected component labels under strict computational constraints?}

This main question decomposes into three fundamental sub-questions:

\begin{enumerate}
    \item \textbf{Self-supervised Landmark Learning}: How can self-supervised methods detect and encode textual or structural component labels as stable landmarks in repetitive tunnel environments?
    
    \item \textbf{Resource Efficiency}: How can SLAM pipelines be adapted for embedded, radiation-tolerant hardware while maintaining real-time operation and mapping accuracy?
    
    \item \textbf{Robustness in Monotonous Geometry}: How can drift and localization ambiguity be reduced in long, GPS-denied tunnels with few natural loop closures?
\end{enumerate}

\subsection{Contribution and Scope of this Work}

Addressing the third sub-question -- robustness in monotonous geometry -- requires fundamental knowledge of odometry error characteristics in accelerator tunnel environments. Without characterizing how different odometry sources accumulate drift under tunnel-specific conditions, it is impossible to derive requirements for SLAM architectures, landmark spacing, loop closure strategies, or sensor fusion approaches. Therefore, \textbf{this paper addresses the prerequisite step}: characterizing odometry error sources through comparative evaluation of wheel odometry, 2D LiDAR odometry, and Deep Visual Stereo Odometry (DVSO) in accelerator tunnel conditions. By quantifying translational and rotational drift behavior, evaluating DVSO as potential alternative or complement to MARWIN's current system, this work establishes the foundational understanding necessary for informed design of the complete resource-efficient, self-supervised SLAM solution outlined in the main research question.

\section{Deep Visual Stereo Odometry (DVSO)} \label{deh:dvso}

Deep visual stereo odometry (DVSO) is a combination of neural networks and classical algorithms that learn features of the world based on moving stereo camera images in a self-supervised manner. These networks can predict optical flow, image depth, visual odometry, and the detection of self-moving objects.  By utilising optical flow and optical flow that can be computed from image depth and visual odometry, DVSO can predict the next images. It compares image predictions with real images using photometric consistency to train itself. To enforce the realistic movement and structure of the 3D world, training is guided by a 3D consistency loss, connecting the predictions of optical flow, image depth and visual odometry to ensure they all correlate with the same underlying 3D world. Although the DVSO was developed to map hard-to-access natural environments, its ability to be self-supervised means it can easily be retrained for use in other environments, such as particle accelerator tunnels. 
Details of the DVSO aproach can be found in \cite{ZAC2025}. 

\section{Experimental Setup} \label{deh:experiment}

To characterize odometry error behavior in accelerator tunnel environments, we conducted a comparative evaluation of wheel odometry, 2D LiDAR odometry, and Deep Visual Stereo Odometry (DVSO) at the European XFEL facility.

\subsection{Test Environment}

\begin{figure*}[!htb]
\centering
\begin{minipage}[t]{0.685\textwidth}
\centering
\includegraphics[width=\textwidth]{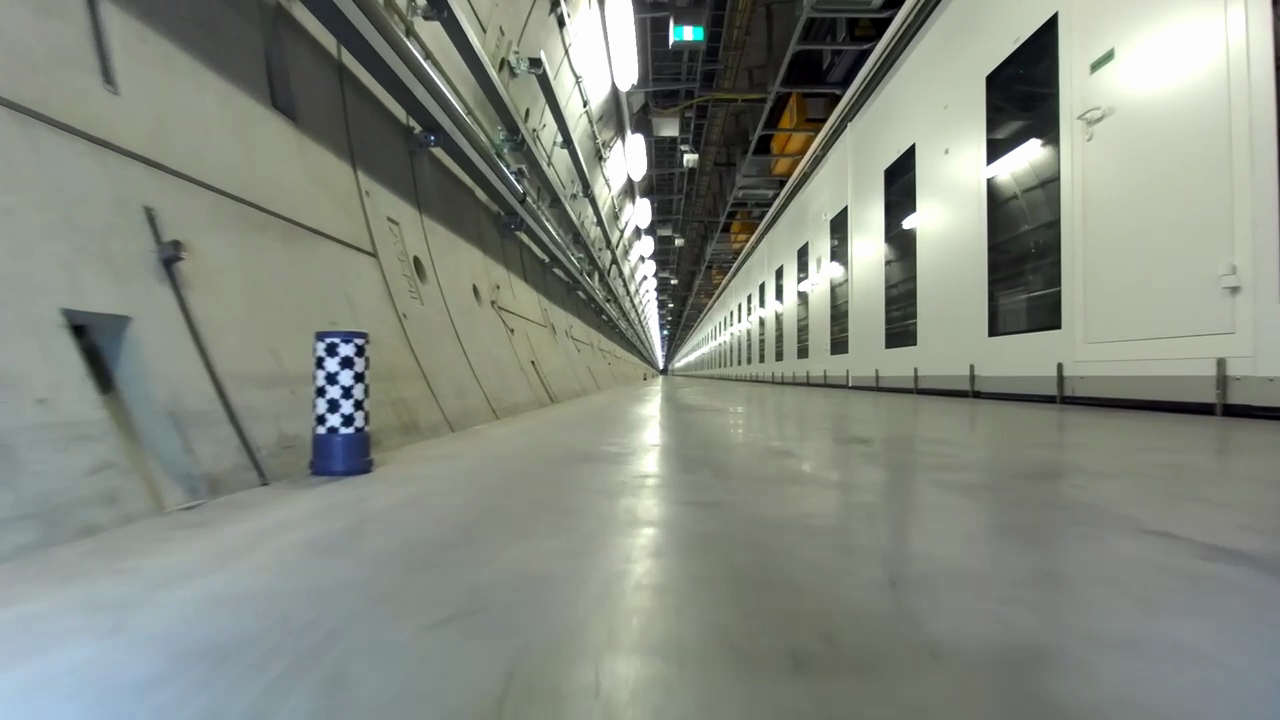}
\caption{Monotonous tunnel section at European XFEL used for odometry evaluation}
\label{deh:tunnel_section}
\end{minipage}
\hfill
\begin{minipage}[t]{0.275\textwidth}
\centering
\includegraphics[width=\textwidth]{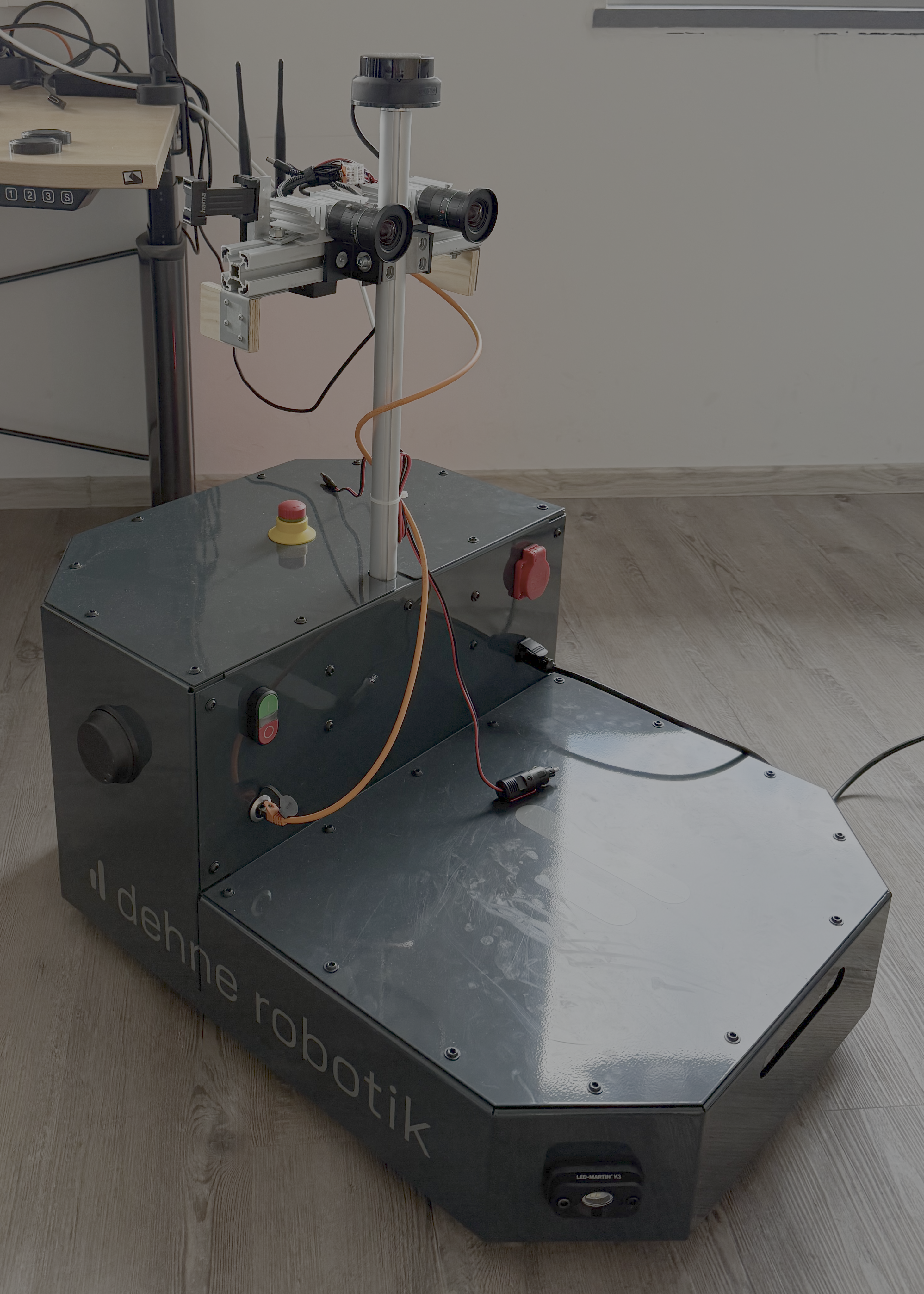}
\caption{Mobile robot platform}
\label{deh:robot_setup}
\end{minipage}
\end{figure*}

The experiment was performed in a particularly monotonous section of the XFEL accelerator tunnel, exemplifying the challenges identified in Section~\ref{deh:challenges}: long straight corridors with repetitive geometric structure, uniform low-texture surfaces, and minimal distinctive visual features (Figure~\ref{deh:tunnel_section}). This section provides representative conditions for evaluating odometry performance under the constraints of accelerator tunnel navigation.

\subsection{Robot Platform and Sensor Configuration}

Due to ongoing modifications on the MARWIN robot, the experiment was conducted using an alternative mobile platform equipped with equivalent sensor modalities. The platform features:

\begin{itemize}
    \item \textbf{Wheel odometry sensors}: Providing proprioceptive motion estimation
    \item \textbf{2D LiDAR scanner}: Enabling geometric feature-based odometry (comparable to MARWIN's current navigation system)
    \item \textbf{Stereo camera system}: Two high-resolution Basler cameras mounted for DVSO data generation
\end{itemize}

The robot configuration with mounted stereo camera system is shown in Figure~\ref{deh:robot_setup}. This sensor suite enables direct comparison of the three odometry sources under identical environmental conditions and trajectory profiles.

\subsection{Trajectory and Data Collection}

The robot was remotely controlled along a straight trajectory of approximately 100 meters, followed by an in-place rotation and return to the starting position. This trajectory profile was selected to (1) accumulate translational and rotational drift over extended straight-line motion (typical for tunnel navigation), (2) isolate rotational error during the turnaround maneuver, and (3) enable closed-loop evaluation by returning to the known starting position. During the trajectory execution, odometry data was simultaneously recorded from all three sources (wheel odometry, 2D LiDAR, stereo cameras for DVSO), ensuring temporal synchronization for comparative analysis.

\subsection{Ground Truth Validation}

To validate the odometry measurements, we deployed four cylindrical markers at 18-meter intervals along the trajectory. Each marker consists of a PuzzlePole \cite{ZAC2025b} -- a cylinder wrapped with the PuzzleBoard calibration pattern \cite{STE2024} (Figure~\ref{deh:tunnel_section}). The PuzzleBoard pattern enables 6-DOF pose estimation, providing high-precision reference positions for odometry error quantification. Comparative evaluation of odometry sources requires quantification of translational and rotational errors accumulated over the trajectory. The PuzzlePoles deployed at surveyed positions along the trajectory serve as reliable ground truth references. Each PuzzlePole detection provides a pose measurement relative to the robot frame, extracted from the stereo camera stream during passage and timestamped for temporal association with odometry measurements. The three odometry sources operate at different frame rates: 2D LiDAR odometry at 10 Hz, wheel odometry at 50 Hz, and DVSO at 5 Hz. As mentioned above all sensors are time-synchronized, allowing interpolation and alignment of odometry frames with landmark observations. The synchronized multi-rate data forms the input to the graph optimization framework, where PuzzlePole detections constrain the odometry trajectories and enable quantification of translational and rotational errors for each odometry source.


\section{Graph-Based Validation} \label{deh:graphvalidation}

To quantify odometry accuracy, we employ graph-based optimization that aligns trajectory estimates with PuzzlePole landmark observations. While PuzzlePoles provide precise pose measurements, they remain sparse compared to the hundreds or thousands of accumulated odometry frames between consecutive landmarks. Direct drift measurement between landmarks proves unsuitable, as errors at trajectory start accumulate significantly while identical errors near the end have minimal impact. Instead, we construct a pose graph where nodes represent robot poses and edges encode odometry transformations. PuzzlePole detections add constraint edges between corresponding nodes. Graph optimization then computes minimal corrections to odometry measurements required for consistency between trajectory and landmarks. The magnitude of these corrections quantifies odometry error and provides a robust validation metric that accounts for error accumulation throughout the trajectory.

\subsection{2D LiDAR Odometry Limitations}

The 2D LiDAR odometry was excluded from the analysis due to fundamental limitations in the tunnel environment. The obstacle-free straight passage provides insufficient geometric features perpendicular to the scan plane, preventing reliable estimation of translational motion along the tunnel direction. This confirms the challenge of monotonous geometry identified in Section~\ref{deh:challenges} and motivates the evaluation of alternative odometry sources.

\subsection{Deep Visual Stereo Odometry Performance}

\begin{figure}[htbp]
\centering
\begin{minipage}{0.49\textwidth}
    \centering
    \includegraphics[width=\textwidth]{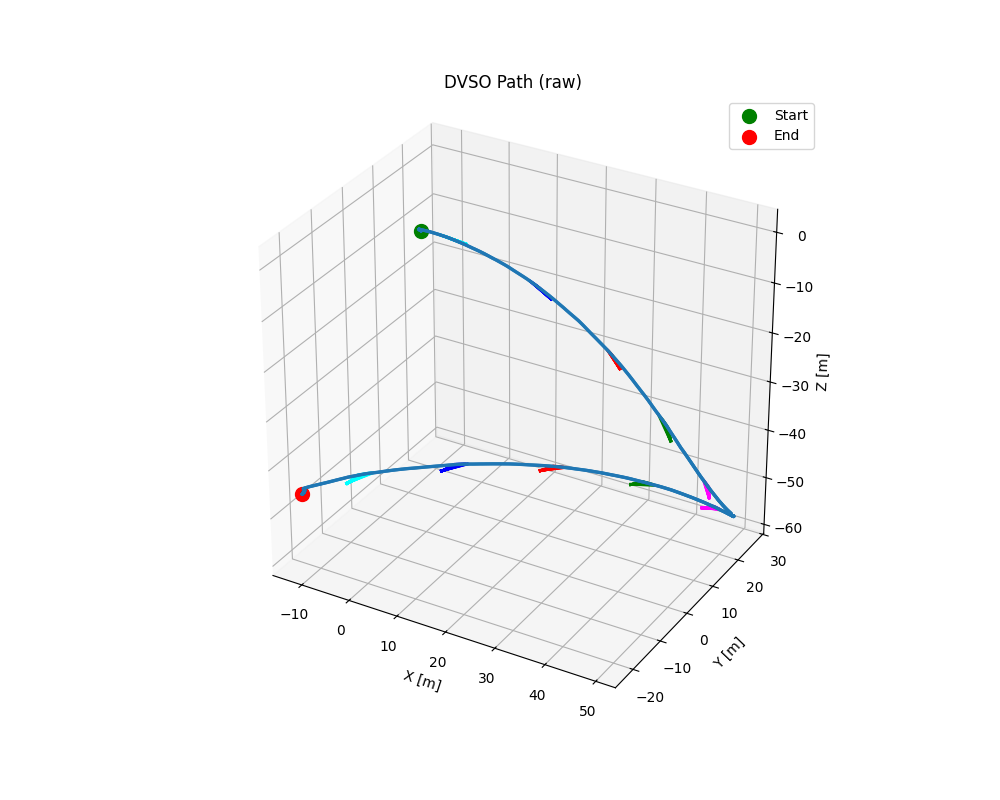}
\end{minipage}
\hfill
\begin{minipage}{0.49\textwidth}
    \centering
    \includegraphics[width=\textwidth]{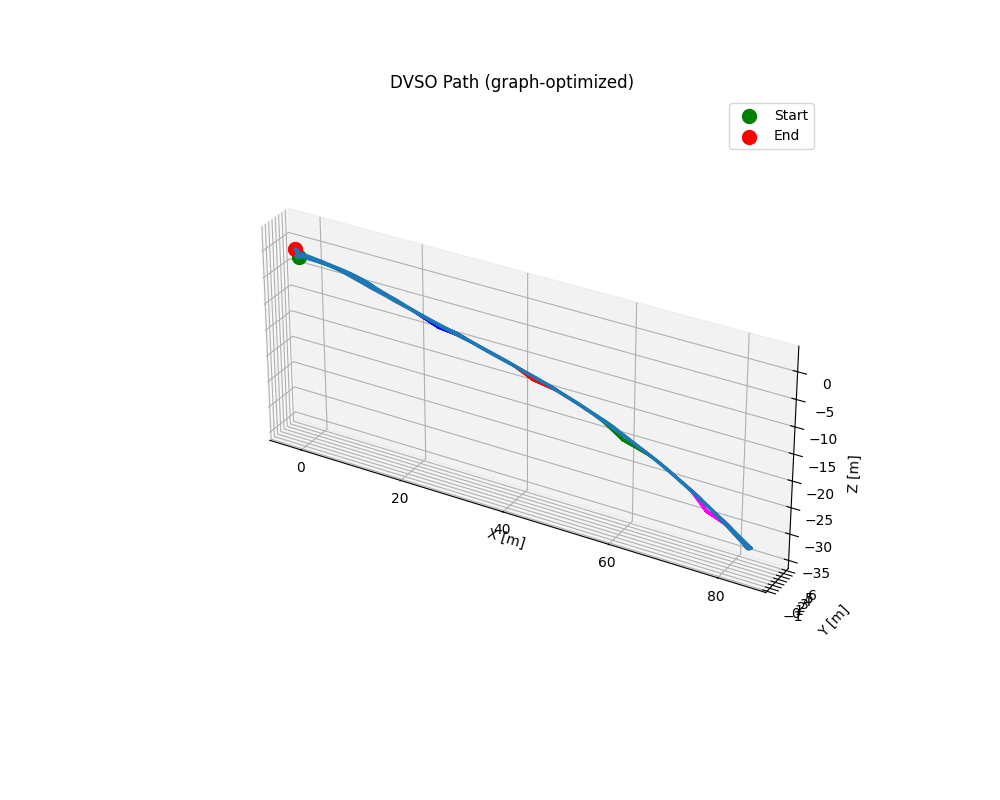}
\end{minipage}
\caption{DVSO trajectory comparison: raw estimates (left) showing downward/rotational drift, and graph-optimized trajectory (right)}
\label{deh:dvso_comparison}
\end{figure}

Figure~\ref{deh:dvso_comparison} presents the DVSO trajectory evaluation. The raw DVSO estimates (left) exhibit a characteristic downward drift shortly after the starting point, potentially due to insufficient training of the underlying neural network on environments matching the planar tunnel geometry. Notably, the trajectory shows considerable separation between start and end points despite the closed-loop nature of the trajectory, suggesting that DVSO experiences difficulties with rotational motion estimation -- particularly during the turnaround maneuver at the far end of the tunnel. The colored markers in the plot indicate the detected PuzzlePole landmarks along the trajectory. The graph-optimized result (right) demonstrates substantial error reduction: the trajectory is constrained by the PuzzlePole observations, significantly minimizing rotational errors and achieving proper closure of the loop.

\subsection{Wheel Odometry Performance}

\begin{figure*}[htbp]
\centering
\begin{minipage}{0.49\textwidth}
    \centering
    \includegraphics[width=\textwidth]{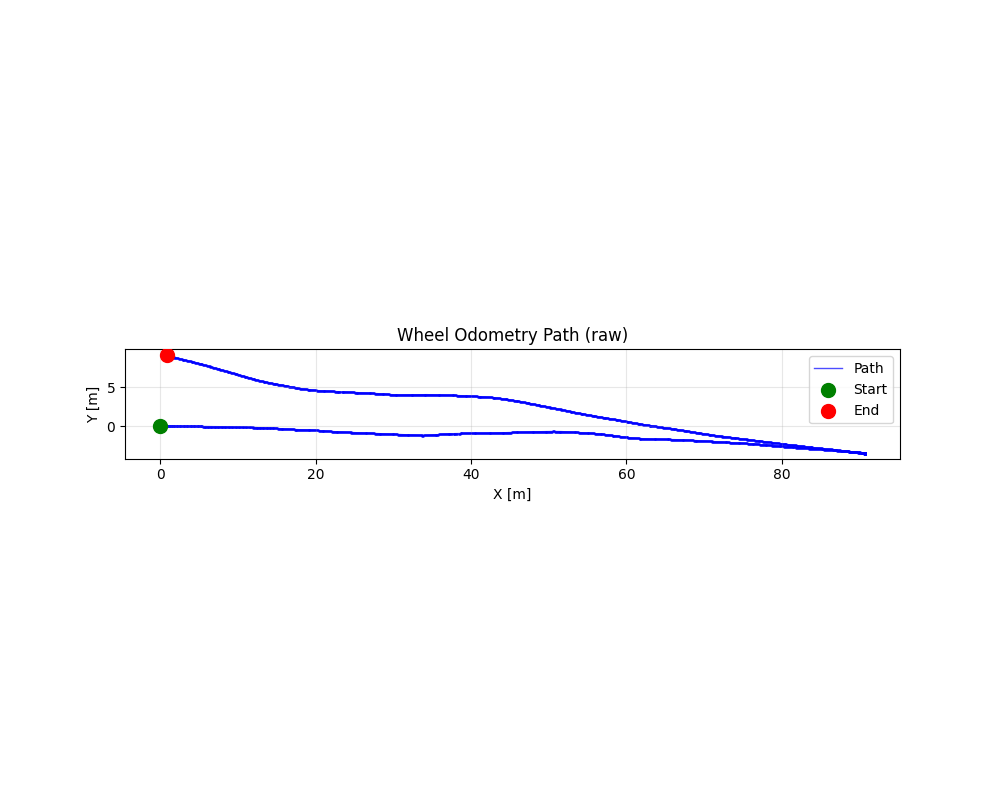}
\end{minipage}
\hfill
\begin{minipage}{0.49\textwidth}
    \centering
    \includegraphics[width=\textwidth]{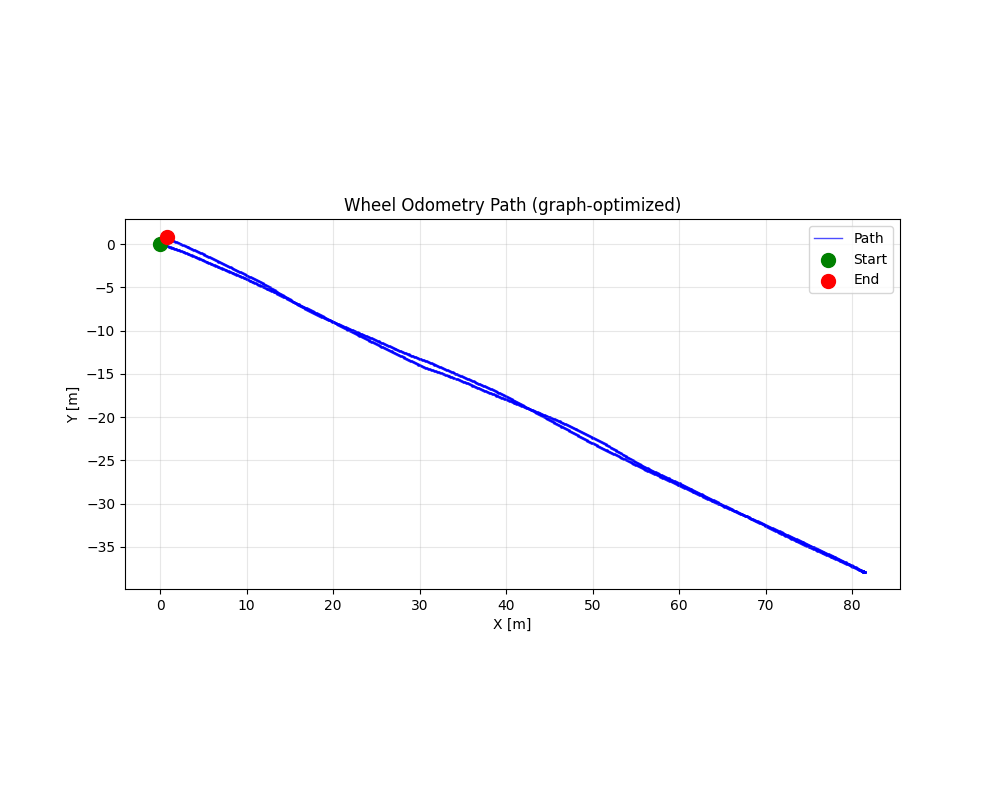}
\end{minipage}
\caption{Wheel odometry trajectory comparison: raw measurements (left) showing cumulative drift, and graph-optimized trajectory (right)}
\label{deh:wheel_comparison}  
\end{figure*}

Figure~\ref{deh:wheel_comparison} shows the wheel odometry trajectory before and after graph optimization. Since wheel odometry provides only 2D pose estimates (x, y, yaw), the graph optimization was constrained to these degrees of freedom, in contrast to the full 3D optimization applied to DVSO data. The raw odometry (left) exhibits cumulative drift over the 100-meter trajectory, though notably the start and end points lie considerably closer together compared to the raw DVSO trajectory, suggesting better rotational consistency during the turnaround maneuver. The graph-optimized trajectory (right) demonstrates successful alignment with the PuzzlePole landmarks. A noticeable offset in the negative y-direction is visible in the optimized result. This offset arises because the graph optimization does not have knowledge of the absolute PuzzlePole positions in the tunnel coordinate frame -- instead, the optimization incorporates only the relative constraints that the markers are spaced 18 meters apart and aligned in a straight line. Despite this global positioning ambiguity, the trajectory alignment relative to the landmark sequence demonstrates effective drift correction through the PuzzlePole constraints.

\subsection{Comparative Analysis}

The graph-based optimization quantifies the average translational and rotational errors for each odometry frame, revealing the correction magnitudes required to align the trajectories with PuzzlePole constraints. Table~\ref{tab:odometry_errors} presents the error characteristics for both odometry sources.

\begin{table}[htbp]
\centering
\caption{Average odometry errors per frame and per second for DVSO and wheel odometry}
\label{tab:odometry_errors}
\vspace{\baselineskip}
\begin{tabular}{lccc}
\hline
\textbf{Odometry Source} & \textbf{Frame Rate} & \textbf{Trans. Error} & \textbf{Rot. Error} \\
\hline
DVSO & 5 Hz & 0.00148 m/frame & 0.043°/frame \\
     &      & \textbf{0.0074 m/s} & \textbf{0.215°/s} \\
\hline
Wheel Odometry & 50 Hz & 0.00018 m/frame & 0.002°/frame \\
               &        & \textbf{0.009 m/s} & \textbf{0.1°/s} \\
\hline
\end{tabular}
\end{table}

When normalized to temporal resolution (errors per second), wheel odometry exhibits comparable translational error accumulation to DVSO (0.009 m/s vs. 0.0074 m/s), despite operating at 10× higher frame rate. However, wheel odometry demonstrates slightly superior rotational accuracy (0.1°/s vs. 0.215°/s), confirming the observation from trajectory analysis that DVSO experiences greater difficulty with rotational motion estimation.

\section{Conclusion} \label{deh:conclusion}

Graph-based validation at the European XFEL facility revealed that 2D LiDAR is inadequate for translational motion estimation in obstacle-free tunnel passages. While this evaluation is based on a single trajectory and requires further validation across diverse tunnel conditions, the results provide initial evidence that DVSO and wheel odometry achieve remarkably similar translational accuracy (0.0074 m/s vs. 0.009 m/s) despite drastically different computational requirements. Given this similarity, deploying DVSO for sensor fusion appears impractical, as long as only visual odometry is used and other prediction like optical flow, 3D world reconstuction, or self moving object detection are not exploited. The substantial computational overhead for stereo processing and neural network inference yields minimal accuracy improvement over encoder-based wheel odometry. For developing a self-supervised SLAM system in accelerator tunnels, these findings suggest prioritizing wheel odometry for translation and computationally cheaper LiDAR-based scans for rotational correction, while semantic landmarks remain essential for periodic drift correction. In future we aim to validate these findings through extended experiments, develop self-supervised landmark learning from text-based infrastructure markers, and address adaptation to radiation-tolerant embedded hardware for autonomous navigation in accelerator tunnels.


\begin{thebibliography}{9}
  
\bibitem{CHE2023}
Chen, C., Liu, X., Li, Y., Ding, L., Feng, C.:
\emph{DeepMapping2: Self-Supervised Large-Scale LiDAR Map Optimization.}
Conference on Computer Vision and Pattern Recognition (CVPR), 2023, pp. 9306--9316, \url{https://doi.org/10.1109/CVPR52729.2023.00898}.

\bibitem{DEH2018}
Dehne, A., Möller, N., Hermes, T.:
\emph{MARWIN: Localization of an Inspection Robot in a Radiation-Exposed Environment.}
Advances in Science, Technology and Engineering Systems Journal, 2018, pp. 354--362, \url{https://doi.org/10.25046/aj030436}.

\bibitem{DIO2022}
Diomede, M., Ayvazyan, V., Branlard,  J., Grecki, M., Hierholzer, M., Hoffmann, M., Lautenschlager, B., Pfeiffer, S., Schmidt, C.,Walker, N.:
\emph{Update on the LLRF Operations Status at the European XFEL.}
arXiv:2210.04711 [physics.acc-ph], 2022, \url{https://doi.org/10.48550/arXiv.2210.04711}

\bibitem{FU2022}
Fu, X., Liu, C., Zhang, C., Sun, Z., Song, Y., Xu, Q., Yuan, X.:
\emph{Self-Supervised Learning of LiDAR Odometry Based on Spherical Projection.}
International Journal of Advanced Robotic Systems, 19(1), 2022, pp. 1--13, \url{https://doi.org/10.1177/17298806221078669}.

\bibitem{PRA2021}
Prados Sesmero, C., Villanueva Lorente, S., Di Castro, M.:
\emph{Graph SLAM Built over Point Clouds Matching for Robot Localization in Tunnels.}
Sensors, 2021, 5340, \url{https://doi.org/10.3390/s21165340}.

\bibitem{STE2024}
Stelldinger, P., Schönherr, N., Biermann, J.:
\emph{PuzzleBoard: A New Camera Calibration Pattern with Position Encoding.}
arXiv:2409.20127 [cs.CV], 2024. \url{https://doi.org/10.48550/arXiv.2409.20127}.

\bibitem{ZAC2025}
Zach, J., Stelldinger, P.:
\emph{Self-Supervised Deep Visual Stereo Odometry with 3D-Geometric Constraints.}
In Proceedings of the 18th ACM International Conference on PErvasive Technologies Related to Assistive Environments (PETRA '25), 2025, pp. 336--342, \url{https://doi.org/10.1145/3733155.3733194}.

\bibitem{ZAC2025b}
Zach, J., Stelldinger, P.:
\emph{PuzzlePoles: Cylindrical Fiducial Markers Based on the PuzzleBoard Pattern.}
Proceedings of the 17th International Conference on Autonomous Systems, 2025.

\bibitem{ZHA2022}
Zhang, R.; Shi, S.; Yi, X.; Jing, M.:
\emph{Application of RGB-D SLAM in 3D Tunnel Reconstruction Based on Superpixel Aided Feature Tracking.}
The International Archives of the Photogrammetry, Remote Sensing and Spatial Information Sciences, XLIII-B2, 2022, pp. 559--564, \url{https://doi.org/10.5194/isprs-archives-XLIII-B2-2022-559-2022}.

\end{thebibliography}
\end{document}